%% file: recsip.tex
\documentclass{svproc}
\usepackage{url}

\usepackage{amsmath,amssymb,amsfonts}
\usepackage{algorithmic}
\usepackage{graphicx}
\usepackage{textcomp}
\usepackage{xcolor}

\usepackage{booktabs}
\usepackage{csquotes}
\usepackage{currfile}
\usepackage{hyperref}
\usepackage{cleveref}
\usepackage{listings}
\usepackage{pgfplotstable}
\usepackage{standalone}
\usepackage{subfiles}
\usepackage{tabularx}
\usepackage{tikz}

\usetikzlibrary{shapes, arrows}

\graphicspath{{figures/}}%

\newcommand{\abs}[1]{\mathinner{\vert#1\vert}}%

\begin{document}
\mainmatter              
\title{RECSIP: REpeated Clustering of Scores Improving the Precision}
\titlerunning{RECSIP}  
%
\author{André Schamschurko \and Nenad Petrovic \and Alois Christian Knoll}
%
\authorrunning{Schamschurko et al.} 
%
\tocauthor{André Schamschurko, Nenad Petrovic and Alois Christian Knoll}
\institute{Technical University of Munich (TUM) \\
School of Computation, Information and Technology (CIT) \\
Chair of Robotics, Artificial Intelligence and Real-time Systems\\
\email{andre.schamschurko@tum.de, nenad.petrovic@tum.de, k@tum.de}}

\maketitle              

\begin{abstract}
The latest research on Large Language Models (LLMs) has demonstrated significant advancement in the field of Natural Language Processing (NLP). However, despite this progress, there is still a lack of reliability in these models.
This is due to the stochastic architecture of LLMs, which presents a challenge for users attempting to ascertain the reliability of a model's response.
These responses may cause serious harm in high-risk environments or expensive failures in industrial contexts.
Therefore, we introduce the framework REpeated Clustering of Scores Improving the Precision (RECSIP) which focuses on improving the precision of LLMs by asking multiple models in parallel, scoring and clustering their responses to ensure a higher reliability on the response.
The evaluation of our reference implementation recsip on the benchmark MMLU-Pro using the models GPT-4o, Claude and Gemini shows an overall increase of 5.8 per cent points compared to the best used model.
\keywords{Large Language Model, Natural Language Processing, RECSIP}
\end{abstract}

\section{Introduction}
\label{sectionIntroduction}
Large Language Models (LLMs) have made significant progress in the field of Natural Language Processing (NLP).
The analysis of their properties and capabilities is part of current research.
The ability of LLMs to interpret and output natural language represents a significant advance with promising applications in industry and other similar use cases.

In the industrial context, LLMs can be used to improve the human-machine interaction.
This includes the control of robots~\cite{10611462} and the planning~\cite{roco2024,zheng2024evaluating} of their actions.
Another popular context is the medical one, using LLMs to help with diagnoses~\cite{zhang_llm-based_2024,medicalUseCase}.

Due to the stochastic architecture of LLMs, it is not possible for the user to trust the responses of LLMs blindly.
In many of the desired use cases of LLMs, their output could potentially cause harm to humans or result in costly failures.
Therefore, in these areas, the accuracy and precision of the answers provided by LLMs are important to increase their reliability.

Our goal in this paper is to increase the truthfulness of LLM responses.
To achieve this, we focus on the precision of LLMs to reduce the number of wrong responses.
Current research focuses on different methods to achieve this goal like multiagent debate (\cref{sectionMultiAgentDebate}), Self-Consistency (\cref{sectionSelfConsistency}) and uncertainty evaluations (\cref{sectionUncertainty}) that come with their own limitations.
In our view, an LLM evaluator (e.g.\ multiagent debate, uncertainty evaluations) limits the reliability of a system, since the truthfulness of the system's responses depends on the truthfulness of the evaluator's LLM.
The reliance of Self-Consistency on a single LLM limits the capabilities of the system to the knowledge and capabilities of that model.
If the model contains outdated data and therefore answers incorrectly, Self-Consistency has no way of recognizing these cases.

Therefore, we approach these limitations by combining the ideas of Self-Consistency~\cite{wang_self-consistency_2023} and multiagent debate~\cite{liang2024encouragingdivergentthinkinglarge}.
We structured our paper as follows: first, we introduce our framework RECSIP, which compares the responses of multiple LLMs to gather different reasoning paths to find a trustful answer and evaluate our implementation \texttt{recsip} on the benchmark MMLU-Pro~\cite{wang2024mmlupro} using the models GPT-4o, Claude and Gemini.
This evaluation shows an overall increase of $5.8$ per cent points to the best used model in \texttt{recsip} and outperforms the leading model in the leaderboard of MMLU-Pro\footnote{\url{https://huggingface.co/spaces/TIGER-Lab/MMLU-Pro}}.

\section{Related Work}
\label{sectionRelatedWork}
Currently, there are multiple approaches to recognize and reduce hallucinations of LLMs.
The approaches differ in the number of language models, in the correction approach and in the scoring of the output to decide which outputs could be hallucinations.
A summary of these can be found in \cref{summaryRelatedWorks}.

\begin{table}
    \caption{Summary of the comparison of related works}
    \label{summaryRelatedWorks}
    \begin{tabularx}{\textwidth}{l|c|X}
        Approach & Multiagent & Description \\
        \hline\hline
        Multiagent debate & yes & A debate between different agents with predefined stances and a judge agent deciding the debate. \\
        \hline
        ReConcile & yes & A debate between different agents with predefined stances and own confidence estimation, which changes over the debate. The debate ends if all confidence scores satisfy a specific threshold. \\
        \hline
        Self-Consistency & no & A decoding technique for CoT generating and comparing multiple reasoning paths of a single LLM. \\
        \hline
        Uncertainty Evaluation & no(yes) & An approach scoring the uncertainty of the output of a LLM on a specific level like sequences or concepts. Many works use a LLM (the same or an additional) as a scorer. \\
    \end{tabularx}
\end{table}

\subsection{Multiagent approaches}
\label{sectionMultiAgentDebate}
The multiagent debate~\cite{liang2024encouragingdivergentthinkinglarge} focuses on debates between different LLM agents which get the other agents output as input.
This approach uses multiple LLMs as debaters and one LLM as judge of the debate.
The debaters have defined roles telling them to be either affirmative or negative debaters.
The goal is to let the debaters express their opinion with their reasoning and then the judge will decide which side of the debate wins.

Chen et al.~\cite{chen-etal-2024-reconcile} introduces ReConcile, a multiagent group discussion with confidence estimation.
Each debater defends and argues for their position finishes their response by adding their own estimation of their confidence.
As they are still receptive to each other's arguments, this confidence score changes during the discussion.
The confidence scores decide when the discussion ends by reaching a specific threshold.

In contrast to these multiagent approaches, RECSIP does not assign a role or opinion to a model.
Instead, RECSIP forwards the same question to all models in parallel and compares their responses.
Another difference is that neither the scoring nor the decision for the final result has to be done by an LLM in RECSIP.

Recent work~\cite{wang2024rethinkingboundsllmreasoning} indicates that a single LLM with a strong prompt can perform comparably to multiagent debates.

\subsection{Self-Consistency}
\label{sectionSelfConsistency}
Self-Consistency~\cite{wang_self-consistency_2023,wan_dynamic_2024,liang_internal_2024,li2024escape,aggarwal2023letssamplestepstep} is a decoding strategy for chain-of-thought (CoT) prompting.
For one CoT prompt, it generates a diverse set of reasoning paths and aggregates all reasoning paths for the final answer.
The idea of this aggregation is that a solution found on multiple reasoning paths has a higher probability of being correct than one with fewer reasoning paths.
Self-Consistency works with only one LLM without any additional verifier or other tool, as it generates the multiple reasoning paths on the decoding step and then evaluates the different answers and paths.

RECSIP reuses the idea of Self-Consistency with multiple reasoning paths by extending it to multiple models and all prompts.

\subsection{Uncertainty Evaluation}
\label{sectionUncertainty}
The field of uncertainty evaluations varies from sequence level uncertainty~\cite{manakul_selfcheckgpt_2023,zheng2024evaluating} to concept level uncertainty~\cite{wang_clue_2024}.
All uncertainty evaluations unite to work on the quantification of the stochastical process of LLMs.
The majority of works in this field employ a second LLM for this purpose or let the LLM quantify its own response.
Each approach annotates the LLM output with scores which are then to be interpreted by the user or a framework to decide how to approach further.

In contrast to these uncertainty evaluations, RECSIP does not quantify the stochastical process of LLMs.
Instead, RECSIP compares the response of a LLM to the ones of other LLMs with the same request and afterwards all models repeatedly choose the most probable answer from a multiple-choice question.

\section{Methodology}
\subsection{Framework}
\begin{figure}
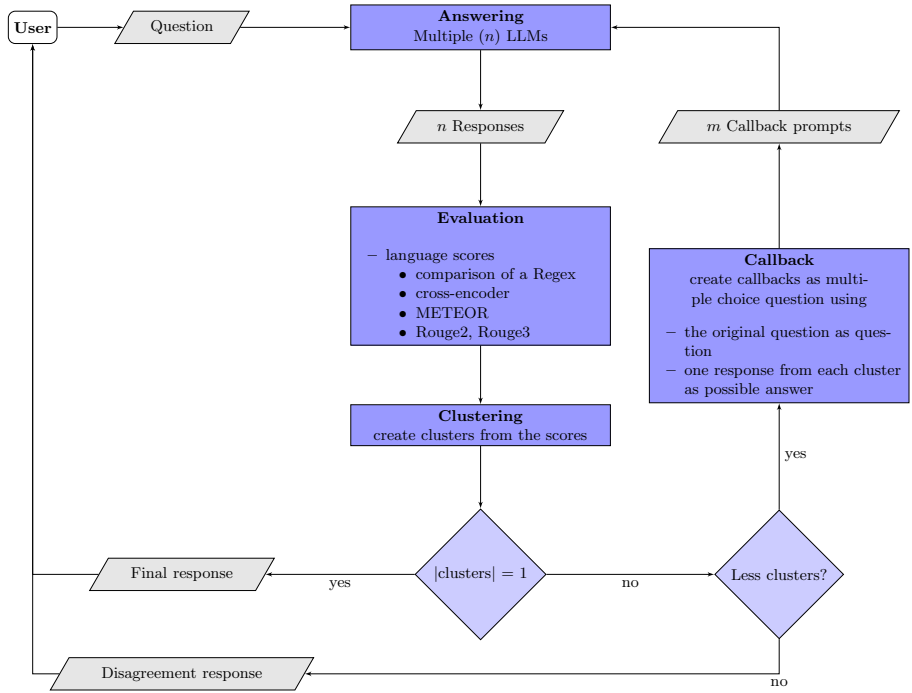

    \centering
    \resizebox{0.975\textwidth}{!}{\subfile{"./figures/implementation/implementation.tex"}}
    \caption{Schema of the RECSIP implementation}
    \label{ImplementationSchema}
\end{figure}

The goal of our proposed framework is to be more confident with the answers of language models which means we want to be sure that when we receive a response, it should be correct in the current context.
Otherwise, there should be either no response or a hint that the answer is not as reliable as wanted.
As the approaches using Self-Consistency~\cite{wang_self-consistency_2023,wan_dynamic_2024,liang_internal_2024} mention, it is a common idea that a solution with many different ways of solving is probably correct.
We propose to apply this idea not by using one LLM with many paths like Self-Consistency does, but by using various LLMs and comparing their reasoning.
As the LLMs are trained on different data sets and some even have a slightly different architecture, it is expected that we receive more different reasoning paths this way.
If the responses of the different LLMs are not similar, they probably have different solutions and, therefore, should take a look at all provided solutions to pick the most probable answer.
In \cref{ImplementationSchema}, we can see how the RECSIP framework works.

Generally, the framework works as follows.
First, we prompt our question to a number of language models, in this example $n$.
Each of them answers the question separately.
Then, we evaluate the $n$ answers of the language models and score their similarity.
After that, we cluster the responses using their scores and count the number of clusters.
If there is more than one cluster, then some responses are different from the others.
Afterwards, we check whether we are able to decide on a correct response out of all $n$ responses.
If not, then we do a callback question to check which of these responses is more probable to reduce the number of different answers and, therefore, reduce the number of clusters.
After the language models answer this callback prompt, we repeat this loop until we are able to decide for a correct answer out of all $n$ responses or the number of clusters does not change any more.
This would mean that the callback prompt did not reduce the number of different responses any more, and therefore, we can expect the LLMs to stick to their responses in future questions.
In this case, we can break the loop and notice the user that the language models did not find an agreement which means that we do not have a trustful answer.

\subsection{Implementation}
The framework RECSIP is very flexible in the steps of the evaluation and the callback.
It allows evaluation implementations from using scores like BLEU~\cite{BLEUscore} to ones using sentence transformer and cross-encoder.
The~\cref{ImplementationSchema} shows the schema of our implementation \texttt{recsip}.

We use the scores ROUGE~\cite{ROUGEscore} and METEOR~\cite{banerjee-lavie-2005-meteor} in combination with a cross-encoder.
ROUGE and METEOR are both scores used in the field of machine translations and are used in our implementation for their simplicity because of the n-gram usage and the ability to ignore synonyms.
Additionally, we use the cross-encoder \texttt{stsb-distil\-roberta-base} to rank the classification loss between the different responses to recognize content-wise similar responses.
For special purposes, we also have a direct comparison using a regular expression match.
This can be used to compare responses where the user cares only about a small part of it, e.g.\ \enquote{The answer is .*\$} for multiple-choice questions.

We do not use a dedicated clustering algorithm like k-means clustering because our scores are relative values between two responses and not absolute scores.
Therefore, our clustering algorithm checks the scores between the responses and creates clusters for binary similar responses.
The needed limit for the similarity between two responses is score dependent.
The regular expressions have to be completely equal.
The cross-encoder score is the most important score, as it encodes the classification loss between the responses which helps us to estimate their relatedness.
However, the cross-encoder penalizes big length differences.
Therefore, ROUGE and METEOR are used to recognize a response as part of the other response by using a limit of $1$ to counter the cross-encoder penalty for length differences.

After clustering, we check whether all LLMs agree which means that we found the correct response.
Otherwise, we check whether we reduced the number of clusters in the last step.
This would mean that at least one possible response got removed from the pool of responses.
Consistency in the number of clusters means that not enough LLMs switched their response in comparison to the previous callback prompt.
Therefore, we will break the loop and forward the question back to the user with the feedback that the LLMs could not agree.

If we had a reduction of the set of clusters to a set with more than one cluster, we create a callback prompt.
For the callback prompt, we construct a multiple-choice question using the original question, picking one random response from each cluster and formulate them as possible answers for the original question.
We can pick the response randomly from each cluster, as we assume all responses in a cluster to be similar.
This multiple-choice question is sent to each LLM separately and asks them to pick the best fitting response for the original question.

\section{Evaluation}
\subsection{Experimental design}
In order to test the effect of RECSIP on the precision of the output, we evaluate our implementation \texttt{recsip} on the benchmark MMLU-Pro~\cite{wang2024mmlupro}.
MMLU-Pro is extending the mostly knowledge-driven benchmark MMLU~\cite{hendrycks2021measuring} by challenging reasoning tasks and adding additional response options so that MMLU-Pro has ten response options instead of four.
These changes lead to a more difficult benchmark for LLMs which is recognizable in the drop of the benchmark scores between MMLU and MMLU-Pro of at least $0.16$.

Given the plug-and-play ability of RECSIP, we can freely pick models for our evaluation.
We pick three high-scoring LLMs from different organizations for our evaluation of \texttt{recsip}: ChatGPT (GPT-4o), Claude (Claude-3.5-Sonnet) and Gemini (Gemini-1.5-Pro).
To ensure the comparability to the scores of the single LLMs measured by the TIGER AI Lab, we use one of their recent snapshots: gpt-4o-2024-08-06, claude-3-5-sonnet-20241022, gemini-1.5-pro-002.

\subsection{Results}
The \cref{PrecisionTable} illustrates the outcome of our benchmarking.
The overall result demonstrates that our implementation using the three models outperforms all models in the MMLU-Pro leaderboard by at least $5.8$ per cent points.
The table demonstrates also that the algorithm of comparing and clustering the responses of the models repeatedly results in an enhanced precision of the LLMs.
The results for the single models are distributed by the TIGER AI Lab on the MMLU-Pro leaderboard and their GitHub repository.
The highest-scoring model of the three LLMs in each category of MMLU-Pro is outperformed by our implementation by a margin ranging from $0.020$ to $0.103$.

\begin{table*}
    \centering
    \caption{Precision comparison of the single models and recsip}
    \resizebox{\textwidth}{!}{\subfile{"./figures/evaluation/precision/table.tex"}}
    \label{PrecisionTable}
\end{table*}

Another interesting finding of~\cref{PrecisionTable} and the leaderboard of MMLU-Pro are the different strengths of the models.
Independent of the fact which model is the best within a specific category, \texttt{RECSIP} outperforms all three models.
In terms of the models used, Gemini is the best model in Physics while ChatGPT is the best model in Biology and Claude in Economics.
An approach for future research would be the analysis of the impact of weighing the responses of the models according to their specific strengths.
This reduces the impact of models in their weak categories but also adds another point of failure as the categorization could fail.

The performance difference is mostly reached by filtering results where the models do not find an agreement.
For these results, our implementation returns \enquote{The models could not agree.} and the last response to the initial question of each model connected to their name sorted by their corresponding clusters.
These results can be interpreted as not known, therefore the user should either reformulate the prompt or use another method than LLMs to obtain the result.
Another step by the user can be to use the pool of responses to find a suitable answer by checking which model responded how.
The user could have a priority list of which model is closer to a suitable response, or perhaps you can determine that specific clusters can be ignored to reduce the number of possible questions.
Afterwards, the user could formulate a question suggesting an answer out of a smaller set of responses.

\begin{table}
    \centering
    \caption{Distribution of the answer type by recsip}
    \subfile{"./figures/evaluation/recsip/recsip-table.tex"}
    \label{DistributionTable}
\end{table}

The~\cref{DistributionTable} presents the distribution of the three different answer types of RECSIP.
These show, that our implementation filters more results in categories that have worse scores with the single models, e.g.\ $0.47$ and $0.38$ in the categories Law and Engineering.
At the same time, \texttt{recsip} filters fewer results where the single models have better results like Biology and Psychology with $0.19$ and $0.17$.

These findings show that RECSIP is an effective method to reduce the number of wrong responses with the cost of a third answer type that gives the responsibility back to the user.
In cases of disagreement, the user is still presented with the proposed responses from each model.
They have the option of either repeating the initial question with an improved prompt, performing the task manually, or attempting to determine the most accurate answer from the proposed ones.

\subsection{Wrong response analysis}
\label{sectionResponseAnalysis}
\begin{figure}
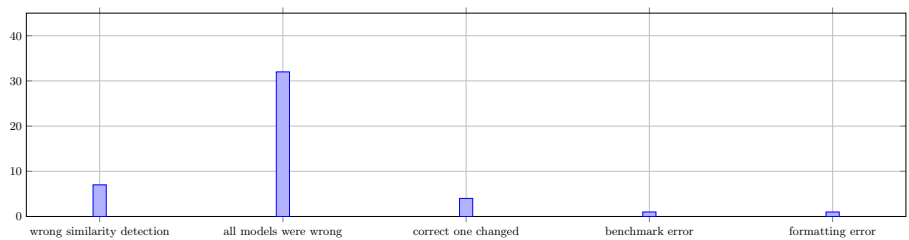

    \centering
    \resizebox{\textwidth}{!}{\subfile{"./figures/evaluation/wrong-responses/diagram.tex"}}
    \caption{Distribution of the reason for wrong responses in Biology}
    \label{wrongResponseDistributionBiology}
\end{figure}

As mentioned in the motivation section, we aim for a reduction of wrong responses from LLMs by applying our methodology.
Therefore, we now analyse the type of reasons for wrong responses by \texttt{recsip}.
The~\cref{wrongResponseDistributionBiology} presents the distribution for these different reasons.
We were able to determine five reasons for wrong responses in the category Biology.

In most cases ($32$ out of $45$), all three models answered wrongly which led to an agreement on a wrong answer as they have to agree on one of the initial answers in RECSIP.
Therefore, RECSIP is reliant on the performance of the single models here, as it can not prevent these cases.

The second most occurring reason ($7$ out of $45$) was the wrong detection of similarity between responses with different answers.
In six of these seven cases, there agreed two models with each other, while a third one picked a different response.
The separate third response was scored by our cross-encoder slightly above our threshold in relation to one of the other two, while the score in relation to the second one was below our threshold.
These cases could be prevented by checking all binary relations to elements of a cluster, before adding responses to an existing cluster.
In the seventh of these cases, this approach would not have been beneficial, since all binary scores exceeded our threshold slightly.
This means we could only increase our threshold to avoid this case and reduce the number of similar cases.
This would probably also reduce the number of right answers and increase the number of disagreements, as correct answers with responses slightly above the threshold would also be affected.
Another approach for this problem would be to improve the evaluation of similarity between the responses, by adding additional scores or improving the scorer.

In four cases, exactly one model was initially correct but switched later to a wrong option which also the others preferred.
These cases are caused by the probabilistic nature of the LLMs and are difficult to avoid completely.
In future research, one approach to reduce the number of occurrences would be prompt engineering to improve the written reasoning of the models for their responses.
With a better and deeper reasoning for their choices, the decisions for the callback question should get more reliable than before.
Another approach could be to weigh the responses of the different LLMs depending on their strength and weaknesses.
This could avoid agreement on responses by models in their weak categories.

\begin{figure}
    \begin{lstlisting}
Certainly! Here are the answers formatted according to your request:
    1. Arthropods Question: 
       - B) The answer is (B).

    2. Hardy-Weinberg Question: 
       - D) The answer is (D).

    3. Mutation and Enzyme Question: 
       - D) The answer is (D).

    4. Recombinant DNA Question: 
       - A) The answer is (A).

    5. Cell Division Control Question: 
       - H) The answer is (H).

    6. Normal Distribution Probability Question: 
       - E) The answer is (E).
    \end{lstlisting}
    \caption{\texttt{recsip} response wrongly interpreted by the benchmark as \texttt{B}}
    \label{wronglyInterpretedResponse}
\end{figure}

In~\cref{wronglyInterpretedResponse}, there we can see the one wrong answer caused by the format interpretation of the benchmark, where the response first answered the five giving examples and answered the main question afterwards.
The benchmark is looking for the first match of \enquote{answer is \textbackslash(?([A-J])\textbackslash)?}, which was the answer of \texttt{recsip} to the first question, in this case \texttt{B}.
Therefore, the benchmark did not find the corresponding answer for the main question and the correct response of GPT-4o was ruled as wrong.

In the last case, the benchmark was looking for the wrong answer.
The question $3011$ asked for the genetic basis of Down's syndrome and expects the answer \enquote{Translocation between chromosomes 14 and 21}.
This describes a rare type of Down's syndrome caused by a novo Robertsonian translocation~\cite{translocationDownSyndromPaper} and not the most common one.
Therefore, this question was answered correctly by \texttt{recsip} saying \enquote{Trisomy of chromosome 21}.

\subsection{Limitations}
\label{sectionLimitations}
One limitation of this work is that our method requires more time and computational power than a question posed to a single LLM.
This is because multiple models must engage with the system at least once with an additional scoring overhead, which introduces additional overheads in terms of processing time and resources.
Therefore, it is necessary to evaluate the cost-value ratio of this overhead on a case-by-case basis depending on the needed reliability of the responses.
Moreover, our framework depends on the performance of the single models, as they are the ones bringing correct solutions into the option pool from which all models pick.
The recognition of cases where all LLMs in RECSIP are wrong remains a future challenge.
There are also cases where the models with initially correct responses switch later to an incorrect one.
These cases are also model specific, but can be addressed by improving the models and by prompt engineering improving the reasoning for a response.

\section{Discussion}
We have shown that the combination of the ideas of Self-Consistency and multiagent debate reduces the number of wrong responses by the system.
Approaching the limitations of RECSIP shown in~\cref{sectionLimitations}, should be part of future research.
Based on our results, the benefits achieved show strong potential for integrating LLM models into automated toolchains.
This would enable trustworthy toolchains with as little human interaction as possible.
Thanks to the modular structure of our methodology, we are able to adapt the comparison score for the different responses according to the application.
We will work on the evaluation of the integration of RECSIP for different tasks in the automated design and software development like requirement analysis~\cite{rag4VehicleDesign}, model-based engineering~\cite{petrovic2025llmbasediterativeapproachmetamodeling}~\cite{petrovic2025multimodalsummarizationmodelbasedengineering} and OCL generation~\cite{oclGeneration}.

In~\cref{sectionResponseAnalysis}, we presented the causes for incorrect response by our system for one specific category.
We showed that there are cases where all LLMs are incorrect in the beginning and agree on one of these responses.
These cases are not detectable by RECSIP by principle and need further research.
We expect the combination with other hallucination reduction methodologies like uncertainty evaluations and retrievable external data like knowledge graphs to be worthwhile.

Further research should also explore different scores for the comparison and evaluation of LLM responses.
Other methodologies, as we have shown in~\cref{sectionRelatedWork}, use either the same LLM or an additional one for scoring.
The use of LLMs to score themselves or other LLMs should be carefully considered in terms of the trustworthiness of the scoring LLM.
In this paper, we avoided this by focusing on n-gram based language scores like ROUGE and METEOR, but we also used a cross-encoder which uses a BERT architecture.
We recommend the exploration of additional scoring architectures and their combinations.
This will also be useful for evaluating output other than natural language, such as code generation or formal languages.

\section{Conclusion}
This paper proposes RECSIP, a novel approach to reduce the number of erroneous responses applicable to all LLMs, and presents the effectiveness of this method.
RECSIP outperforms not only all single performances of its used models, but also demonstrates the capacity to surpass the best single models on the benchmark MMLU-Pro with three models outside the top three.

The analysis of possible reasons for incorrect responses by our implementation \texttt{recsip} suggests that there are cases where RECSIP can not prevent returning an incorrect response.
For those cases, RECSIP is dependent on the models themselves and additional methods to increase the precision of LLMs.

These combinations with additional methods should be included in future work.
Furthermore, we outline prospective research pathways for the advancement of RECSIP, including weighing the LLM responses according to their strength and possible improvements for the similarity scoring and clustering of \texttt{recsip}.

\section{Acknowledgment}
This research was funded by the Federal Ministry of Education and Research of Germany (BMBF) as part of the CeCaS project, FKZ: 16ME0800K.

%
%

\bibliographystyle{spmpsci}
\bibliography{source/literature.bib}

\end{document}

%% file: figures/implementation/implementation.tex
\tikzstyle{connector} = [draw, -latex']
\tikzstyle{data}=[trapezium, draw, text centered, trapezium left angle=60, trapezium right angle=120, minimum height=2em, fill=gray!20]
\tikzstyle{database}=[cylinder, aspect=.5, draw, text centered, trapezium left angle=60, trapezium right angle=120, minimum height=2em, fill=gray!80]
\tikzstyle{decision} = [diamond, draw, text centered, minimum height=2em, fill=blue!20]
\tikzstyle{process} = [rectangle, draw, text centered, text width=5cm, minimum height=2em, fill=blue!40]
\tikzstyle{terminator} = [rectangle, draw, text centered, rounded corners, minimum height=2em]

\hspace{-1cm}
\begin{tikzpicture}
    \node [terminator] at (-5,-4) (user) {\textbf{User}};
    \node [data] at (-2, -4) (question) {Question};
    \node [process] at (4, -4) (llms) {\textbf{Answering}\\Multiple ($n$) LLMs};
    \node [data] at (4, -6) (responses) {$n$ Responses};
    \node [process] at (4, -9) (evaluation) {\textbf{Evaluation}\\
        \begin{itemize}
            \item language scores
            \begin{itemize}
                \item comparison of a Regex
                \item cross-encoder
                \item METEOR
                \item Rouge2, Rouge3
            \end{itemize}
        \end{itemize}};
    \node [process] at (4, -12) (cluster) {\textbf{Clustering}\\
    create clusters from the scores};
    \node [decision] at (4, -15) (score) {$\abs{\text{clusters}}$ = 1};
    \node [decision] at (10, -15) (clusterCheck) {Less clusters?};
    
    \node [data] at (10, -6) (prompts) {$m$ Callback prompts};
    \node [process] at (10, -10) (callback) {\textbf{Callback}\\create callbacks as multiple choice question using
    \begin{itemize}
        \item the original question as question
        \item one response from each cluster as possible answer
    \end{itemize}};
    
    \node [data] at (-2, -15) (response) {Final response};
    \node [data] at (-2, -17) (disagreementResponse) {Disagreement response};

    \path [connector] (user) -- (question);
    \path [connector] (question) -- (llms);
    \path [connector] (llms) -- (responses);
    \path [connector] (responses) -- (evaluation);
    \path [connector] (evaluation) -- (cluster);
    \path [connector] (cluster) -- (score);
    \path [connector] (score) -- node[anchor=north] {yes} (response);
    \path [connector] (response) -| (user);

    \path [connector] (score) -- node[anchor=north] {no} (clusterCheck);
    \path [connector] (clusterCheck) -- node[anchor=west] {yes} (callback);
    \path [connector] (callback) -- (prompts);
    \path [connector] (prompts) |- (llms);

    \path [connector] (clusterCheck.south) |- node[anchor=north] {no} (disagreementResponse);
    \path [connector] (disagreementResponse) -| (user);
\end{tikzpicture}

%% file: figures/evaluation/precision/table.tex
\pgfplotstableset{
    create on use/increase/.style={
        create col/expr={abs(\thisrow{recsip}) - max(\thisrow{gemini-1.5-pro-002}, \thisrow{gpt-4o-2024-08-06}, \thisrow{claude-3-5-sonnet-20241022})}}
}

\pgfplotstabletypeset[
    columns={{ }, gemini-1.5-pro-002, gpt-4o-2024-08-06, claude-3-5-sonnet-20241022, recsip, increase},
    columns/ /.style={string type, column type/.add = {}{|}},
	columns/gemini-1.5-pro-002/.style={column type/.add = {}{|}},
    columns/gpt-4o-2024-08-06/.style={column type/.add = {}{|}},
    columns/claude-3-5-sonnet-20241022/.style={column type/.add = {}{|}},
    columns/recsip/.style={column type/.add = {}{||}},
    fixed, fixed zerofill, precision = 3,
    every head row/.style = {
        before row=\bottomrule,after row=\toprule\bottomrule},
    every last row/.style={before row=\toprule\bottomrule, after row = \toprule}
]{\currfiledir gemini-gpto4-claude-recsip-precision.dat}

%% file: figures/evaluation/recsip/recsip-table.tex
\pgfplotstableset{
    create on use/share of disunity/.style={
        create col/expr={\thisrow{disunity} / (\thisrow{correct} + \thisrow{wrong} + \thisrow{disunity})}}
}

\pgfplotstabletypeset[
    columns={{ },correct,wrong,disunity,share of disunity},
    columns/ /.style={string type, column type/.add={}{|}},
    columns/disunity/.style={column type/.add={}{||}},
    columns/share of disunity/.style={fixed, fixed zerofill, precision = 2},
    every head row/.style={
        before row=\bottomrule,after row=\toprule\bottomrule},
    every last row/.style={before row=\toprule\bottomrule, after row=\toprule}
]{\currfiledir recsip.dat}

%% file: figures/evaluation/wrong-responses/diagram.tex
\pgfplotstableread{\currfiledir wrong-responses.dat}\datatable

\def\xlistmacro{}
\def\xliststring{}
\pgfplotstableforeachcolumnelement{error type}\of\datatable\as\entry{%
\xifinlist{\entry}{\xlistmacro}{}{
        \listxadd{\xlistmacro}{\entry}
        \edef\xliststring{\xliststring\entry,}
    }
}
\hspace{-2.25cm}
\begin{tikzpicture}
\begin{axis}[
    xtick = data,
    xticklabel style={align=center},
    enlarge x limits = true,
    symbolic x coords/.expand once = {\xliststring},
    ybar, ymin = 0, ymax = 45,
    grid = both,
    height = 7cm,
    width = 25cm,
    legend pos = south east
]
\addplot table[y = error number] from \datatable ;
\end{axis}
\end{tikzpicture}